\documentclass[a4paper,font=10]{jpconf}
\usepackage{graphicx} 
\usepackage{iopams}
\usepackage{amsmath}
\usepackage{hyperref}

\newtheorem{prop}{Proposition}
\def\ie{{\it i.e.}}

\def\cf{{\it cf.}}

\begin{document}

\hspace*{\fill}  TIF-UNIMI-2023-8

\title{Product Jacobi-Theta Boltzmann machines with score matching}
\author{Andrea Pasquale$^{1,2}$, Daniel Krefl$^{3}$, Stefano Carrazza$^{1,2,4}$ and\\ Frank Nielsen$^{5}$}

\address{$1$ TIF Lab, Dipartimento di Fisica, Universit\`a degli Studi di Milano and INFN Sezione di Milano Via Celoria 16, 20133, Milan, Italy.}
\address{$2$ Quantum Research Center, Technology Innovation Institute, Abu Dhabi, UAE.}
\address{$3$ Department of Computational Biology, University of Lausanne, Switzerland}
\address{$4$ CERN, Theoretical Physics Department, CH-1211, Geneva 23, Switzerland.}
\address{$5$ Sony Computer Science Laboratories Inc, Tokyo, Japan}

\ead{andrea.pasquale@unimi.it, daniel.krefl@unil.ch, stefano.carrazza@unimi.it, frank.nielsen.x@gmail.com}


\begin{abstract}
The estimation of probability density functions is a non trivial task 
that over the last years has been tackled with machine learning techniques. Successful applications can be obtained using models inspired by the Boltzmann machine (BM) architecture.
In this manuscript, the product Jacobi-Theta Boltzmann machine (pJTBM) is introduced as a restricted version of the Riemann-Theta Boltzmann machine (RTBM) with diagonal hidden sector connection matrix. We show that score matching, based on the Fisher divergence, can be used to fit probability densities with the pJTBM more efficiently than with the original RTBM.
\end{abstract}

\section{Introduction}
The modelling of general probability density functions (PDFs) is a difficult task, even in the one-dimensional setting. Perhaps most commonly applied for this purpose is the method of kernel density estimation (KDE) \cite{kde}. Though often KDE yields a relatively good approximation to the underlying density in small dimensions, its general modelling capacity is limited as only the type of kernel and hyperparameter bandwidth are optimizable degrees of freedom.   

More recently, significant success in modelling arbitrary probability densities, in particular in the high dimensional setting, has been achieved with new models like \emph{variational autoencoders} (VAE)~\cite{vae} and \emph{normalizing flows} \cite{nflow,NormalizingFlow-2020}. Though these models shine in their modelling capacity and ability to easily draw samples from  the modelled densities, obtaining derived quantities like moments, cumulative density functions, conditionals or marginalizations, is not straight-forward.

Unrelated to the above developments, another novel method has been proposed in recent years to model arbitrary probability densities \cite{2020}. The method is based on a generalization of the well-known Boltzmann machine \cite{hinton1983}, dubbed Riemann-Theta Boltzmann machine~\cite{2020} (RTBM). The RTBM features a novel parametric density involving the Riemann-Theta (RT) function, which possesses a significant modelling capacity. The analytic expression for the density allows to explicitly calculate derived quantities. However, one drawback of the RTBM, which hindered so far a wider usage, is that it is computationally expensive, in particular for high dimensional hidden state spaces. The reason being that the computational burden of the RT function scales exponentially with the dimension $d$.

As a step to overcome this computational challenge, we introduce in this work a simplified version of the RTBM by making use of a factorizing property of the RT function. This simplified version, which we will refer to as {\it product Jacobi-Theta Boltzmann machine}, is very suitable for a more efficient optimization (training) using the score matching method of \cite{score}. 

As we will demonstrate experimentally on two and three-dimensional examples, this simplified model yields a significant improvement over the original RTBM based modelling. 

\subsection{Riemann-Theta Boltzmann machine}
The Riemann-Theta Boltzmann machine (RTBM) is a novel variant of a Boltzmann machine, introduced in \cite{2020}. The RTBM is defined
by the following energy model:
\begin{equation}
\label{model}
    E(x) = \frac{1}{2} x^{t} A x + B^t x \,\quad {\rm with}\quad x = \left( \begin{array}{c}
h \\
v
\end{array} \right), \
 B = \left( \begin{array}{c}
B_h \\
B_v
\end{array} \right), \ 
A = 
\left( \begin{array}{cc}
Q & W^t \\
W & T\end{array} \right),
\end{equation}
where $A$ is a positive-definite $N \times N$ matrix and $B$ a $N$ dimensional bias vector.
 For reasons becoming more clear below, $x$, $B$ and $A$ are decomposed into a visible sector $v$ of dimension $N_v$ and a hidden sector $h$ of dimension $N_h$ so that $N = N_v + N_h$. In detail, $Q$ corresponds to the connection matrix of the hidden sector, $T$ to the connection matrix of the visible sector and $W$ encodes the coupling between the two sectors. The Boltzmann distribution for this model is defined as
$
P(v, h) = \frac{e^{-E(v,h)}}{Z} $, where $Z$ denotes the canonical partition function.

The peculiarity of the RTBM is that the hidden
state space is taken to be $\mathbb{Z}^{N_h}$, while the visible state space is $\mathbb{R}^{N_v}$. This quantization of the hidden state space makes the computation of the canonical partition function tractable. In particular, by marginalizing out the hidden sector, a closed form analytic expression for the visible sector probability density function, denoted as $P(v)$, can be derived as \cite{2020}
\begin{equation}
    \label{prob v}
    P(v) = \sqrt{\frac{\det{T}}{(2\pi)^{N_v}}} e^{- \frac{1}{2}\big( 
    ( v + T^{-1}B_v)^t T ( v + T^{-1}B_v)  
   \big)}
    \frac{\tilde{\theta}(B^t_h+v^t W \vert Q)}
    {\tilde{\theta}(B^t_h-B_v^t T^{-1} W \vert Q - W^t T^{-1} W)} \ ,
\end{equation}
where $\tilde{\theta}$ is given by the RT function \cite{book} with rescaled arguments, \ie,
\begin{equation}
    \label{theta1}
     \tilde{\theta}(z \vert \Omega) := \theta\bigg( \frac{z}{2 \pi i} \Big\vert \frac{i \Omega}{2 \pi} \bigg) = \sum_{n \in \mathbb{Z}^k} e^{-\frac{1}{2}n^t \Omega n + n^t z} \,.
    \end{equation}

\subsection{Product Jacobi-Theta Boltzmann machine}
\begin{figure}[h]
\begin{minipage}[b]{0.45\textwidth}
    \includegraphics[width=\textwidth]{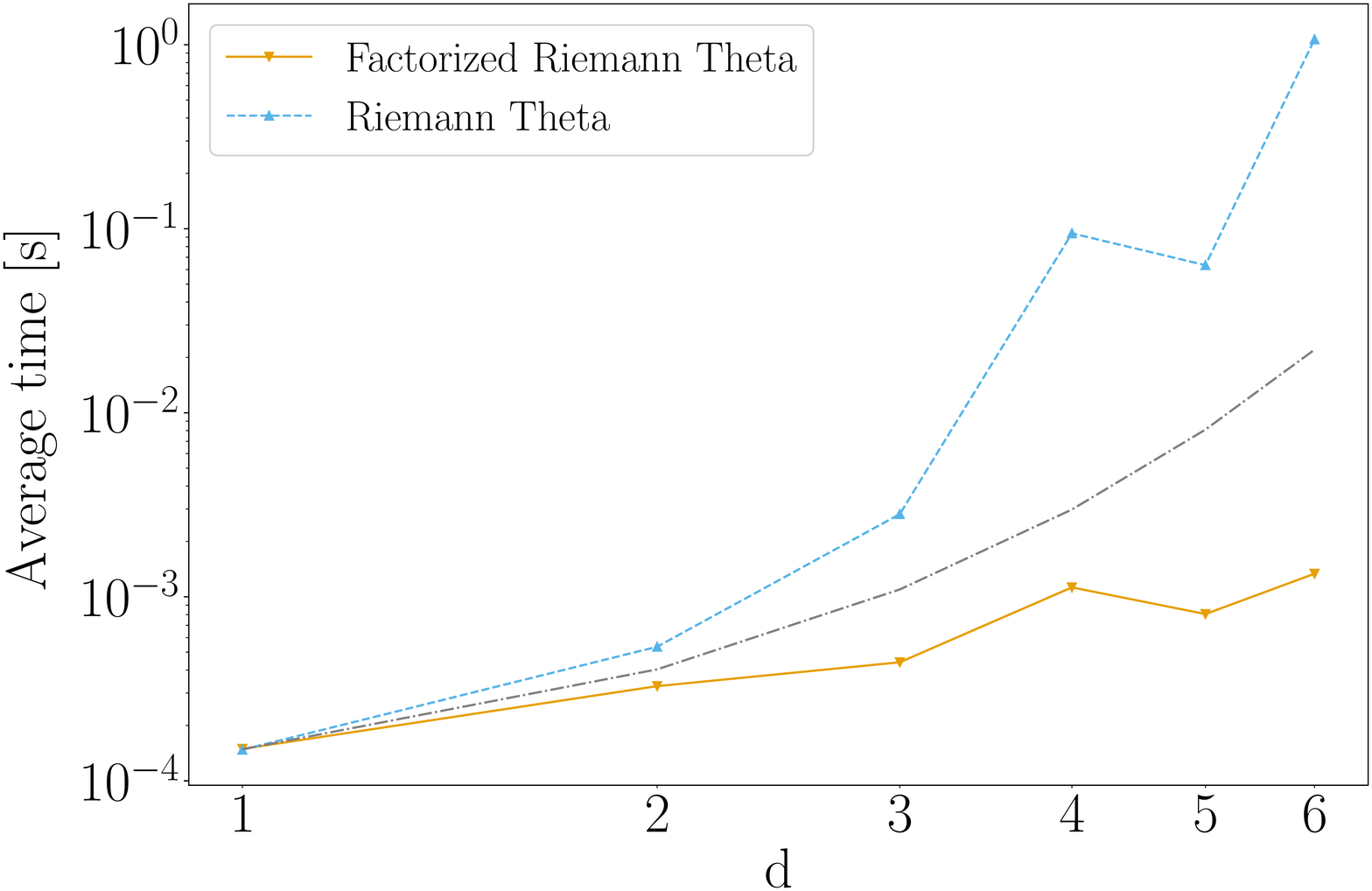}\hspace{2pc}%
\end{minipage}
\begin{minipage}[b]{0.55 \textwidth}
\raggedleft
\caption{\label{fig:factorized} \footnotesize Average time to evaluate the RT function \cite{openRT} for different dimensions $d$ of a diagonal matrix 
        $\Omega$ using the factorized form (yellow curve) and the standard form (light-blue curve). Note that both axes are plotted in logarithmic scale. 
        The matrix elements of $\Omega$ have been sampled
        uniformly from the imaginary unit interval, and we averaged over 10 independent runs.
        Exponential growth is marked with a gray dashed-dotted line. A linear regression in log-log space implies that the average
        time for the factorized RT grows as $\propto d^{1.2}$ with $R^2 = 0.93$.}
\end{minipage}

\end{figure}

The general $\theta$-function factorizes for a diagonal second argument matrix $\Omega$ into one-dimensional $\theta$-functions, which are known as Jacobi-Theta functions: 
$
    {\theta}(z \vert \Omega) = \prod_i {\theta}(z_i \vert (\Omega)_{ii})\,.
$
Consequently, $\tilde{\theta}$ also  factorizes similarly. This factorization has significant computational advantages, as the computational complexity of the non-factorized $\theta$ grows exponentially with the dimension of $\Omega$. For illustration, we plotted in Fig.~\ref{fig:factorized} a run-time comparison between the factorized and non-factorized calculation of $\theta$ with diagonal $\Omega$: While for small $d$ the factorized computation run-time grows experimentally as $d^{1.2}$, the non-factorized shows clearly exponential growth under increasing $k$.


A diagonal $Q$ in $P(v)$ corresponds to a hidden sector which is not inter-connected, and therefore is similar in spirit to the well-known restricted Boltzmann machine, \cf, \cite{ACKLEY1985147}. However, for a RTBM with diagonal $Q$, which we will refer to as {\it product Jacobi-Theta Boltzmann Machine} (pJTBM), the $\theta$-function occurring in the normalization of $P(v)$ does in general not factorize, as that would require that $W$ diagonalizes $T^{-1}$. Hence, at first sight the computational advantage of a pJTBM is limited, as the normalization would still be computationally expensive.

\section{The Fisher cost function}
\label{sec:Fisher}
\subsection{Introduction}
In the original work introducing the RTBM \cite{2020} it has been shown that the probability density of the visible sector $P(v)$ can be used to approximate the underlying density of a given dataset via maximum likelihood estimation (MLE).
  
The novel contribution of this work consists in adopting a different learning method for the RTBM known as \emph{score matching}, first introduced in \cite{score}. Score matching is particularly efficient for non-normalized models, \ie, models
where the partition function $Z$ cannot be computed either analytically or numerically. For instance, this can be the case for high-dimensional densities where the computation of the partition function requires the calculation of a non-trivial multi-dimensional integral. This parameter learning method can also be exploited for models where the partition function can be computed, such as the RTBM, but is computationally expensive: Avoidance of the computation of the partition function during optimization may strongly accelerate the training process. In particular, this applies to the pJTBM defined above, for which the normalization is the main remaining computational obstacle.

Suppose that we are trying to model the probability density $p(x)$ using a parametric
density $q(x, \theta)$, where $\theta$ are the parameters to be optimized. The divergence metric associated with score matching~\cite{score} reads
\begin{equation}
    D_F(p\lvert \lvert q_\theta) = \int p(x) \Bigg\vert 
    \frac{\nabla_{x} \ p(x)}{p(x)} -
    \frac{\nabla_{x} \ q(x,\theta)}{q(x,\theta)}  \Bigg\vert^2 d x \ , 
\end{equation}
which is often referred to as Fisher divergence or relative Fisher information~\cite{RelativeFI-2000}.
It has been shown \cite{score, lyu2012interpretation} that if $q(x,\theta)$ is
sufficiently regular and differentiable, then we can simplify the above expression via integration by parts.
For a finite sample $v$ of $p(x)$ we obtain (up to an irrelevant constant)
\begin{equation}
\label{fisher cost}
    D_F(p\lvert \lvert q_\theta) \approx \mathcal{C}^F = \sum_{i=1}^N \Bigg \vert 
    \nabla_{v_i} \log{q(v_i, \theta)} \Bigg \vert^2
    + 2 \Delta_{v_i} \log{q(v_i, \theta)}\,,
\end{equation}
where $\nabla$ and $\Delta$ denotes the gradient and Laplacian operators, respectively.
We   refer to $\mathcal{C}^F$ as the Fisher cost function. 

\subsection{Derivation of the Fisher cost function for the RTBM and pJTBM}
If we model the probability density $p(x)$ using a RTBM, we can exploit the fact that we have an analytical expression for $q(v,\theta)$, given by $P(v)$ in Eq.~\ref{prob v}, to compute explicitly the Fisher cost function. The two terms in Eq. \ref{fisher cost} for $\mathcal{C}^F$ are given for the RTBM by
\begin{equation}
 \begin{split}
  \partial_{v_i} \log{P(v)} = - ( T v)_i - (B_v)_i + ( W D)_i\,,\quad
  \partial^2_{v_i} \log{P(v)} = - T _{ii} + (W H W^t)_{ii} + ( W D)^2_i\,,
\end{split}   
\end{equation}
with $D$ the normalized gradient and $H$ the normalized Hessian: 
\begin{equation}
    (D)_i = \frac{\nabla_i \tilde{\theta}(B^t_h+v^t W \vert Q)}{\tilde{\theta}(B^t_h+v^t W \vert Q)} 
    \,,  \quad
    (H)_{ij} = \frac{\nabla_i \nabla_j \tilde{\theta}(B^t_h+v^t W \vert Q)}{\tilde{\theta}(B^t_h+v^t W \vert Q)}\,.
\end{equation}

The calculation of $D$ and $H$ can be computationally expensive, especially in the case of 
$N_h > 1$. However, for the pJTBM, \ie, a  \emph{restricted} RTBM with diagonal $Q$ matrix, we can simplify the above expressions using the fact that the $\theta$-function (and consequently also
$\tilde{\theta}$) factorizes. 
This yields
\begin{equation}
\begin{split}
     \partial_{v_i} \log{P(v)} &= - ( T v)_i - (B_v)_i + \sum_{j=1}^{N_h} \frac{\partial_{v_i} \tilde{\theta}((B^t_h+v^t W)_j \vert Q_{jj})}{\tilde{\theta}((B^t_h+v^t W)_j \vert Q_{jj})} W_{ji}\,,\\
     \partial_{v_i}^2 \log{P(v)} &= - T_{ii} + 
     \sum_{j=1}^{N_h} \frac{\partial^2_{v_i} \tilde{\theta}((B^t_h+v^t W)_j \vert Q_{jj})}
     {\tilde{\theta}((B^t_h+v^t W)_j \vert Q_{jj})} W_{ji}^2    
     - 
     \sum_{j=1}^{N_h} \frac{(\partial_{v_i} \tilde{\theta}((B^t_h+v^t W)_j \vert Q_{jj}))^2}
     {\tilde{\theta}((B^t_h+v^t W)_j \vert Q_{jj})} W_{ji}\,.
\end{split}
\end{equation}
Note that, the above simplification is in a computational sense. The complexity of the Fisher cost function is now growing linearly with $N_h$, because we have only to compute
derivatives of Jacobi-Theta functions. This motivates us to adopt a dedicated implementation for the Jacobi-Theta function, which is more efficient,  instead of using  the RT function implementation of \cite{openRT}. For more details about the implementation of the Jacobi-Theta function we refer to the Appendix. 

\section{Example}
As an example we employ RTBMs to model the empirical bivariate distribution given by the \texttt{uranium} dataset \cite{uranium}, which is included in the R package \texttt{copula} \cite{Hofert2020-jg}.
We also fit the models to the empirical joint density of daily log returns of three different equities: \texttt{AAPL}, \texttt{MSFT} and \texttt{GOOG}.

We compare the performance of the standard RTBM, \ie, a generic RTBM optimized via maximum likelihood estimation of parameters,
with a pJTBM trained using the Fisher cost function, as described in section \ref{sec:Fisher}.
Note that we make use of the transformation property of the RTBM under affine transformations \cite{Carrazza:2018nmd} to preprocess the data using a z-score normalization and PCA \cite{pca}. In particular, we perform a pre-training step by fitting first the marginals of
the underlying distribution using 1-dimensional RTBM or pJTBM.

At hand of these examples, we want to address two main points: firstly, we aim to quantify the speed-up achieved using this new learning methodology by benchmarking the time
necessary to run 500 iterations using  CMA-ES as optimizer \cite{cma}. Secondly, we want to compare the quality of the fits by using as \emph{Goodness-of-Fit} test a generalized version of the Kolmogorov-Smirnov (KS) test called Fasano-Franceschini test (FF) \cite{FF-1987}. In particular, we are interested to check if the pJTBM can achieve a comparable level of accuracy, despite the fewer parameters due to diagonal $Q$.

Results are given in Table \ref{tab:results}. We observe that the time required to train a pJTBM is exponentially suppressed compared to the RTBM. This enabled us to push the number of hidden nodes up to 8 with acceptable computational times under our used optimization scheme. Note that for the RTBM we could not go beyond $N_h=4$ in reasonable time.

The FF values are in general comparable between RTBM and pJTBM under the error bounds. However, we observe that the variance for the pJTBM is in general larger compared to the RTBM, despite the lower number of parameters. A possible explanation could be that a model with diagonal $Q$ is more rigid and therefore has more difficulties to reach a good solution from a given initial condition.

 We believe that under a different optimization scheme (optimization on manifolds), the pJTBM could become less dependent on initial condition and also more suitable for far higher-dimensional problems, leading to interesting applications. We hope to investigate this aspect in future work.



\begin{table}

\begin{minipage}[t]{0.6 \textwidth}
\begin{tabular}{lllll}
\br
 & $N_h$ & parameters & Execution Time & FF \\
\mr
\hline
pJTBM  & 2 & 13 & 9.2 s  & $2.32\pm 0.23$\\
RTBM    & 2 &  14 & 59.6 s & $1.72 \pm 0.10$\\
\mr
pJTBM   & 4 & 21 & 20.6 s & $1.81 \pm 0.32$ \\
RTBM   &  4& 27 & 5 min  & $1.50 \pm 0.08$  \\
\mr
pJTBM  & 6  & 29 & 51.6 s & $1.63 \pm 0.22$ \\
RTBM   & 6  & 44 & -  \\
\mr
pJTBM  & 8 & 37   & 82.9 s & $1.76 \pm 0.32$ \\
RTBM   & 8 & 65 & -  \\
\mr\hline
pJTBM  & 3 & 24 & 32.2 s  & $2.91 \pm 0.24$\\
RTBM    & 3 & 27 & 6 min & $2.41 \pm 0.43$\\
\br
\end{tabular}
\end{minipage}
\hspace{0.1 \textwidth}
\begin{minipage}[c]{0.3 \textwidth}
\caption{\footnotesize \label{tab:results} Execution time and FF values for RTBM and pJTBM with different number of
hidden nodes $N_h$. Each result is averaged over 10 independent runs. The results with $N_h$ even correspond to fitting the models to the \texttt{uranium} dataset, while for $N_h = 3$ the models are fitted to the joint \texttt{AAPL}, \texttt{MSFT} and
\texttt{GOOG} stock daily log return distribution. The error on the FF corresponds to the standard deviation between 10 runs of the training process. A $-$ indicates that the computation did not finish in reasonable time.}
\end{minipage}
\end{table}

\begin{figure}[h]
\begin{minipage}{0.7 \textwidth}
\includegraphics[width=0.5 \textwidth]{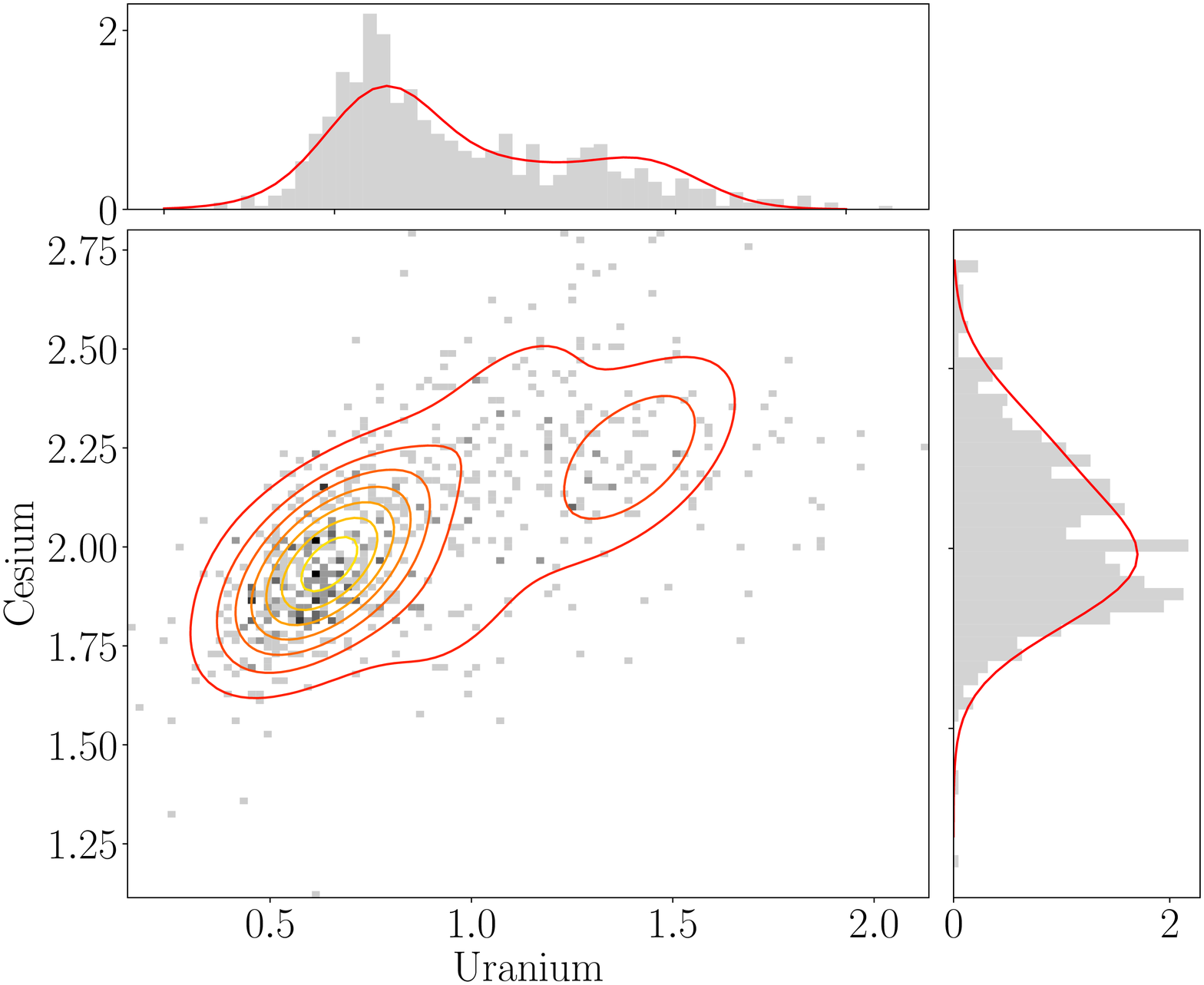}
\includegraphics[width=0.5 \textwidth]{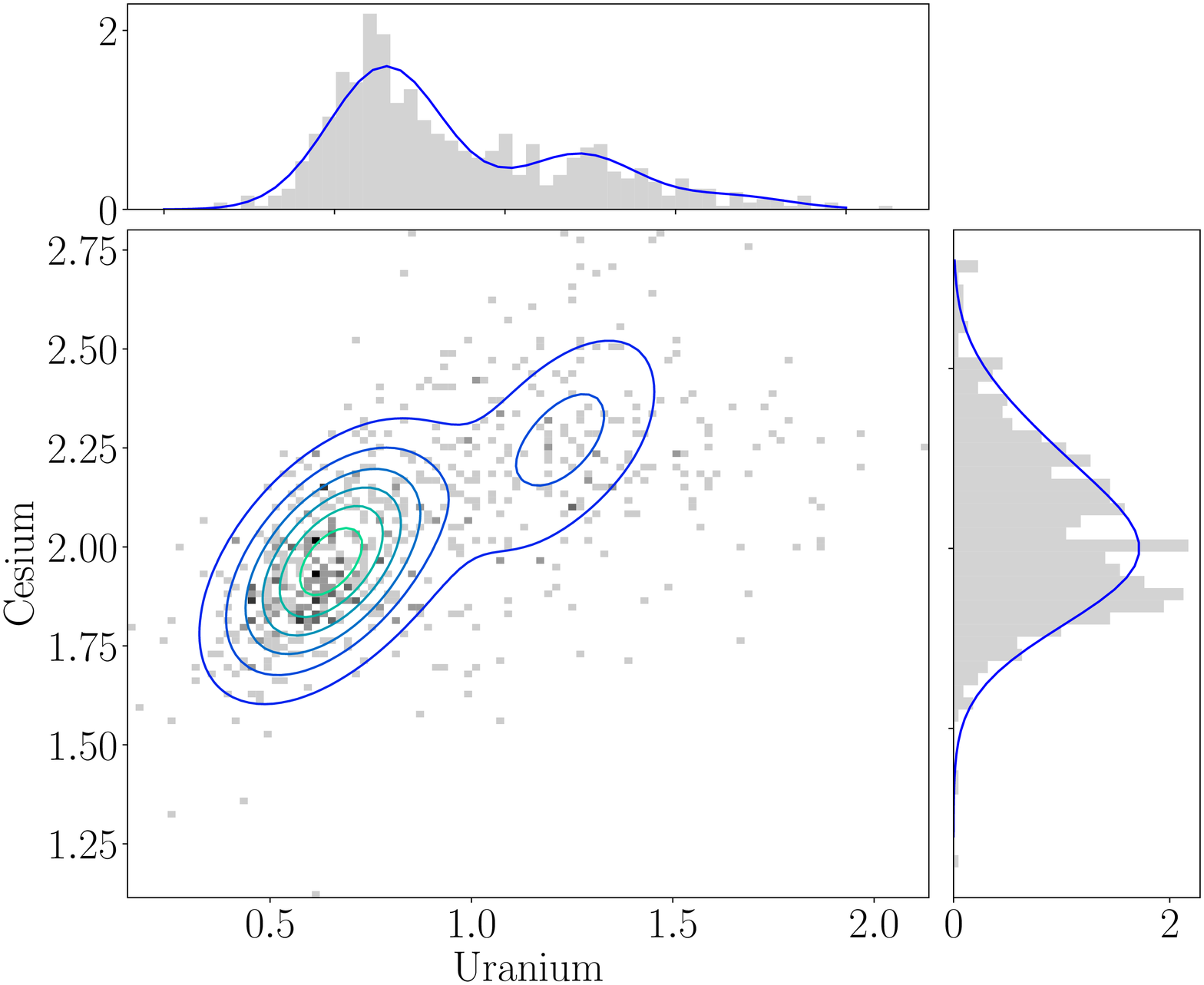}
\end{minipage} 
\hspace{0.05 \textwidth}
\begin{minipage}{0.25 \textwidth}
    \caption{ \footnotesize \label{fig:pJTBM}Contour plots for the pJTBM (left, red) and for the RTBM (right, blue) for the \texttt{uranium} dataset. The $N_h=4$ model with the lowest FF value over 10 independent runs is plotted.}
\end{minipage}
\end{figure}

\appendix
\section*{Appendix}
\setcounter{section}{1}
\label{appendix}
The computation of the RT function and its derivatives is computationally challenging, as the function is defined as an infinite sum over a $N$-dimensional
integer lattice $\mathbb{Z}^N$, \cf, Eq. \ref{theta1}. 
For a general dimension $N$, the RT function and its derivatives can be calculated up to any desired precision $\epsilon$ via summing over a finite subset of lattice points falling inside an ellipsoid of radius $R(\epsilon)$ \cite{deconinck2002computing, SWIERCZEWSKI2016263}.

\begin{figure}[h]
\begin{minipage}[b]{0.45 \textwidth}
\includegraphics[width=\textwidth]{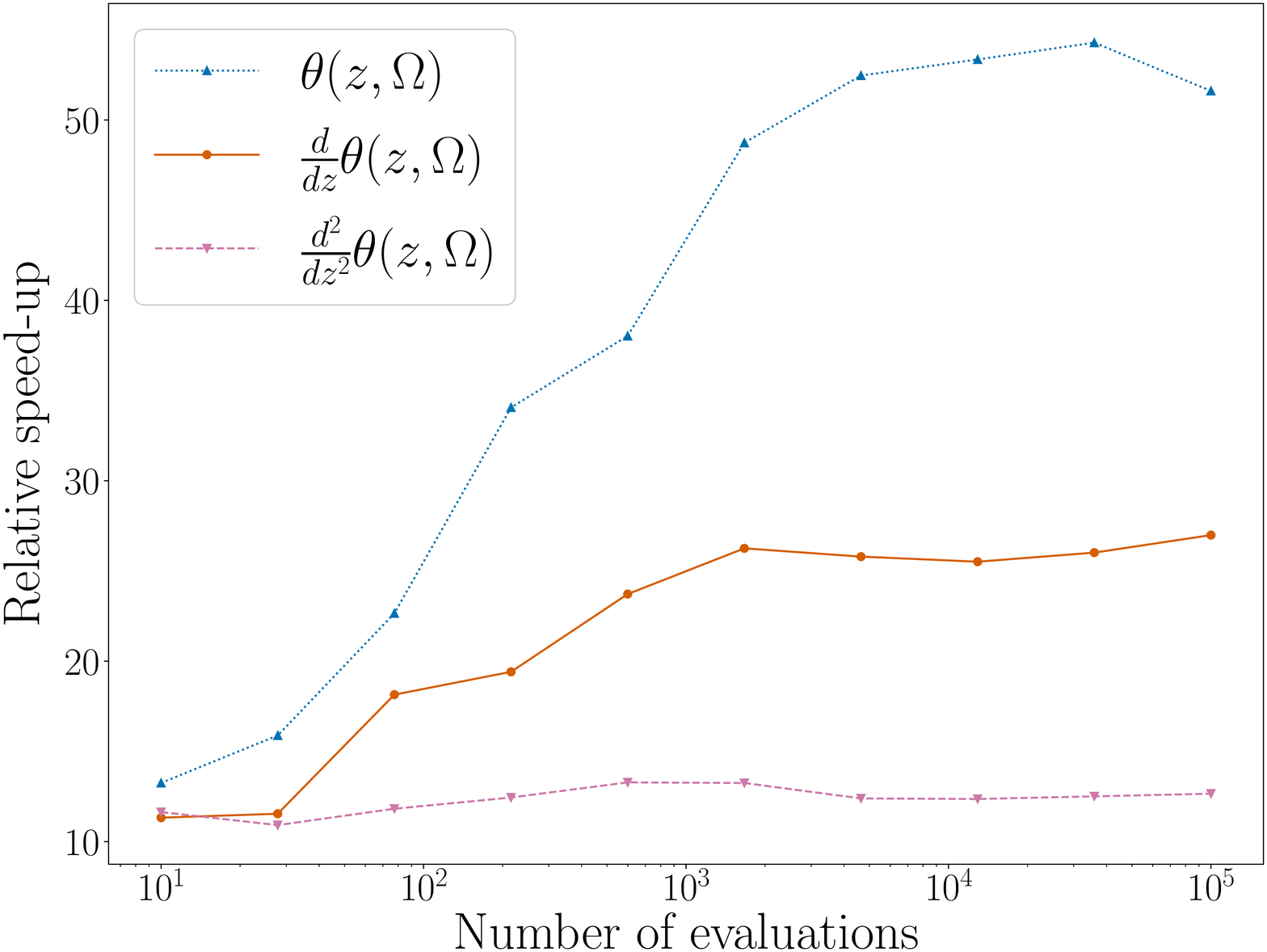}
\end{minipage}
\begin{minipage}[b]{0.55 \textwidth}
 \caption{\label{fig:theta} \footnotesize
 Relative speed-up factor ($t_1 / t_2$)  between the execution time of \cite{deconinck2002computing} ($t_1$),
        and the naive algorithm from \cite{theta} extended to the calculation of the derivatives ($t_2$), as described in the Appendix.
        The parameters of the RT function have been sampled uniformly from the imaginary unit interval and we averaged over 10 independent runs. }
    
\end{minipage}
\end{figure}

For the purpose of this work we mainly need an efficient implementation of the RT function and its derivatives for $N=1$. A highly efficient implementation has been proposed in \cite{theta}, albeit not including the derivatives. For simplicity, we only consider here the naive algorithm given in \cite{theta}, because it can be easily generalized to compute as well the derivatives, and gives us already a significant performance gain compared to \cite{deconinck2002computing}, see  Fig.~\ref{fig:theta}. 

The naive algorithm for evaluation of the 1d $\theta$ function starts from the partial summation
\begin{equation}
    S_B(z, \Omega) = 1 + \sum_{0 < n < B}  q^{n^2} (e^{2 \pi i n z} + e^{- 2 \pi i n z}) =: 
    1 + \sum_{0 < n < B} v_n \ ,
\end{equation}
where $q = e^{i \pi \Omega}$ and $v_n = q^{n^2} (e^{2 \pi i n z} + e^{- 2 \pi i n z}) $.
One can prove that for specific ranges of $\Omega$ and $z$ there exist a $B(\epsilon)$ such that an approximation to the $\theta$ function with absolute precision $\epsilon$ is obtained. Once bounded, $S_B$ can be computed via exploiting the fact that the summands $v_n$ can be computed recursively, \ie,
$
    v_{n+1} = q^{ 2 n} v_1 v_n - q^{4 n} v_{n-1}$.
In the following, we will generalize this recursive algorithm to the computation of the derivatives. We made an implementation of this algorithm available inside the {\it Theta} (v0.0.2) Python package \cite{carrazza_stefano_2017_1120325}.

\subsection{Partial derivatives of the 1d Riemann-Theta function}
The recursive algorithm recalled above can be easily generalized as follows:  First, note that it is not hard to see that the partial sums for the first and second derivative read
\begin{equation}
\begin{split}
     U_B(z, \Omega) &= \sum_{1 < n < B}  - 4 \pi n  q^{n^2} \sin(2 \pi n z) =:  \sum_{1 < n < B} w_n \ ,\\
     V_B(z, \Omega) &= \sum_{1 < n < B}  - 8 \pi^2 n^2 q^{n^2} \cos(2 \pi n z) =:  \sum_{1 < n < B} \xi_n \ .
\end{split}
\end{equation}
The recurrence relations for $w_n$ and $\xi_n$ can be derived quite easily by making use of Chebyshev's polynomials.
This yields
\begin{equation}
    w_{n+1} =  (n+1) \bigg[ \frac{2 \cos(2 \pi z)}{n} q^{2n + 1} w_n - 
                \frac{q^{4n}}{n-1} w_{n-1}\bigg]\,,
\end{equation}
and
\begin{equation}
       \xi_{n+1} = (n+1)^2 \bigg[ \frac{2 \cos( 2 \pi z)}{n^2} q^{2 n + 1} \xi_n 
        - \frac{1}{(n-1)^2}  q^{4n}\xi_{n-1} \bigg]\,.
\end{equation}
It remains to derive the ranges of validity for which the partial sums for $U_B$ and $V_B$
give an accurate approximation for the gradients of the RT function:

\begin{prop}
    Suppose that $\Im(\tau) \ge 0.742$ and $0 \le \Im(z) \le \Im(\tau) /2$.

    Then, for $B \ge 1$, $|\frac{d}{dz} \theta(z,\tau) - U_B(z,\tau) | \le 3|q|^{(B-1)^2}$, where
    \begin{equation}
        U_B(z,\tau) = \sum_{0 < n < B} - 4 \pi n q^{n^2} \sin(2 \pi n z)\,.
    \end{equation}
Proof. \emph{In order to demonstrate the proposition we bound the remainder of the series:}
\begin{equation}
    \begin{split}   
         | \frac{d}{dz}  \theta(z,\tau) - U_B(z,\tau) |
   \le &  
    \ 4 \pi \sum_{n \ge B} |q |^{n^2}  \  
    \frac{n}{2} \big| e^{2 \pi i n z} - e^{- 2 \pi i n z}\big| 
    \le  \ 4 \pi \sum_{n \ge B} n | q| ^{n^2-n}
    \\
    \le & \ 4 \pi \sum_{n \ge B} n | q| ^{(n-1)^2} 
    \le  \ 8 \pi | q| ^{(B-1)^2}\sum_{n \ge 0} n |q|^n
    = \ 8 \pi  \frac{| q| ^{(B-1)^2+1}}{(1 - |q|)^2}\,.
    \end{split}
\end{equation}
\emph{Numerically it can be shown that for $\Im(\tau)\ge 0.742$, 
we have $\frac{8\pi |q|}{(1-|q|)^2} \le 3$, which proves the proposition.} 
\end{prop}

\begin{prop}
    Suppose that $\Im(\tau) \ge 0.882$ and $0 \le \Im(z) \le \Im(\tau) /2$.

    Then, for $B \ge 1$, $| \frac{d^2}{dz^2}\theta(z,\tau) - V_B(z,\tau) | \le 3|q|^{(B-1)^2}$, where
    \begin{equation}
        V_B(z,\tau) = \sum_{0 < n < B} - 8 \pi ^2 n^2 q^{n^2} \cos(2\pi n z)\,.
    \end{equation}
Proof. \emph{We bound the remainder of the series as follow:}
\begin{equation}
    \begin{split}   
         |\frac{d^2}{dz^2}\theta(z,\tau) - V_B(z,\tau) |
   \le &  
    \ 8 \pi^2 \sum_{n \ge B} |q |^{n^2}  \  
    \frac{n^2}{2} | e^{2 \pi i n z} | + |e^{- 2 \pi i n z}|
    \le  \ 8 \pi^2 \sum_{n \ge B} n^2 | q| ^{n^2-n} \\
    \le & \ 8 \pi^2 \sum_{n \ge B} n^2 | q| ^{(n-1)^2} 
    \le  \ 32 \pi^2 | q| ^{(B-1)^2}\sum_{n \ge 0} n^2 |q|^n
    =   \frac{\ 64 \pi^2 | q| ^{(B-1)^2+2}}{(1 - |q|)^3}\,.
    \end{split}
\end{equation}
\emph{Numerically it can be shown that for $\Im(\tau)\ge 0.882$, 
we have $\frac{8\pi |q|^2}{(1-|q|)^3} \le 3$, which proves the proposition.}
\end{prop}

\section*{References}
\bibliographystyle{iopart-num} 
\bibliography{ref} 

\end{document}